\documentclass[preprint,12pt]{elsarticle}

\usepackage{amssymb}
\usepackage{amsmath}
\usepackage{multirow}
\usepackage{hyperref}
\usepackage{url}
\usepackage{graphicx} 
\usepackage{caption} 
\usepackage{subcaption} 
\usepackage{booktabs} 
\usepackage{array} 
\usepackage{xcolor} 
\usepackage{makecell} 
\usepackage{tabularx} 
\usepackage{threeparttable} 
\usepackage{adjustbox} 
\usepackage{tikz} 

\biboptions{square,sort&compress}

\hypersetup{
    colorlinks=true,
    linkcolor=blue,
    filecolor=magenta,
    urlcolor=blue,
    citecolor=blue
}

\makeatletter
\def\ps@pprintTitle{%
     \let\@oddhead\@empty
     \let\@evenhead\@empty
     \def\@oddfoot{}%
     \let\@evenfoot\@oddfoot}
\makeatother

\begin{document}

\begin{frontmatter}

\title{SARD: A Large-Scale Synthetic Arabic OCR Dataset for Book-Style Text Recognition} 

\author[inst1]{Omer Nacar\corref{cor1}} 
\ead{onajar@psu.edu.sa}

\author[inst1]{Yasser Al-Habashi}
\ead{yalhabashi@psu.edu.sa}

\author[inst1]{Serry Sibaee}
\ead{ssibaee@psu.edu.sa}

\author[inst1]{Adel Ammar}
\ead{aammar@psu.edu.sa}

\author[inst1]{Wadii Boulila}
\ead{wboulila@psu.edu.sa}

\affiliation[inst1]{organization={Prince Sultan University},
            city={Riyadh},
            country={Saudi Arabia}}

\cortext[cor1]{Corresponding author}

\begin{abstract}
Arabic Optical Character Recognition (OCR) is essential for converting vast amounts of Arabic print media into digital formats. However, training modern OCR models, especially powerful vision-language models, is hampered by the lack of large, diverse, and well-structured datasets that mimic real-world book layouts. Existing Arabic OCR datasets often focus on isolated words or lines or are limited in scale, typographic variety, or structural complexity found in books. To address this significant gap, we introduce SARD (Large-Scale Synthetic Arabic OCR Dataset). SARD is a massive, synthetically generated dataset specifically designed to simulate book-style documents. It comprises 843,622 document images containing 690 million words, rendered across ten distinct Arabic fonts to ensure broad typographic coverage. Unlike datasets derived from scanned documents, SARD is free from real-world noise and distortions, offering a clean and controlled environment for model training. Its synthetic nature provides unparalleled scalability and allows for precise control over layout and content variation. We detail the dataset's composition and generation process and provide benchmark results for several OCR models, including traditional and deep learning approaches, highlighting the challenges and opportunities presented by this dataset. SARD serves as a valuable resource for developing and evaluating robust OCR and vision-language models capable of processing diverse Arabic book-style texts.\end{abstract}


\end{frontmatter}

\section{Introduction}
The digitization of vast archives of Arabic printed materials, such as books, manuscripts, and historical documents, represents a critical endeavor for preservation, accessibility, and scholarly analysis. A significant portion of these valuable resources exists solely in image formats, rendering them inaccessible to modern text-based tools for search, analysis, and automated processing~\cite{al2020review}. Optical Character Recognition (OCR) serves as the core technology enabling the conversion of these images into machine-readable text. This transformation facilitates a wide array of applications, including the development of digital libraries, enhancement of information retrieval systems, automation of mail handling processes, and improvement of document verification procedures~\cite{Yarmouk}. Nevertheless, accurately recognizing Arabic text poses unique challenges attributable to its cursive script, context-dependent letter shapes, ligatures, and the considerable variation across different fonts and writing styles.

Achieving high accuracy in Arabic OCR, particularly for the complex layouts characteristic of books, necessitates powerful models trained on extensive and diverse datasets. Recent advancements, especially within deep learning and vision-language models, have demonstrated considerable promise by learning rich visual and linguistic features~\cite{najam2023analysis}. The effective training of such models heavily relies on the availability of high-quality, paired image and text data. A comprehensive dataset is crucial for models to master the intricacies of Arabic script variations, manage different font types, and interpret the structural organization of document pages~\cite{mosbah2024adocrnet}.

\begin{table*}[t]
\centering
\caption{Comparison of SARD with existing Arabic OCR datasets.}
\label{tab:ocr_datasets_comp}
\adjustbox{width=\textwidth}{ 
\begin{tabular}{@{}llcccccc@{}} 
\toprule
\textbf{Dataset Name} & \textbf{Year} & \textbf{Type} & \textbf{Nb images} & \textbf{Total Words} & \textbf{Fonts Used} & \textbf{Focus} \\
\midrule
PATS-A01~\cite{PATS} & 2009 & Printed & 22K & 8,248 & 8 & Classic Arabic Literature (Lines) \\
APTI~\cite{mosbah2024adocrnet} & 2013 & Printed (Synthetic) & 45M & 45M & 10 & Single-word OCR \\
UPTI~\cite{sabbour2013segmentation} & 2013 & Printed & 10K & 12K & 1 (Urdu) & Urdu OCR \\
AlexuW~\cite{AlexU-Word} & 2014 & Handwritten & 25K & 10,989 & - & Segmented Letter-based Recognition \\
PAW~\cite{article_PAW} & 2017 & Printed & ~400K & ~550K & 5 & Sub-word OCR \\
MADBase~\cite{ElSawy2016CNNFH} & 2017 & Handwritten & 70K & 700K & - & Char-Based Recognition \\
Yarmouk~\cite{Yarmouk} & 2018 & Printed & ~9K & 436,921 & - & Diverse Printed Text (Pages) \\
OnlineKHATT~\cite{Online-KHATT} & 2018 & Handwritten (Online) & 10K & 22,216 & Multiple & Handwritten Arabic (Words/Lines) \\
Shotor~\cite{asadi2020shotor} & 2020 & Printed (Synthetic) & 120K & 62,900 & Various (Farsi) & Farsi OCR \\
ABAD~\cite{ICDAR} & 2021 & Handwritten & 15K & 14.4M & - & Tunisian Place Names (Words) \\
IDPL-PFOD~\cite{farsitext} & 2021 & Printed (Synthetic) & 30K & 38,476 & Various (Farsi) & Synthetic Farsi OCR (Pages) \\
Arabic-Nougat~\cite{arabicnougats} & 2023 & NN Extracted & 13.7K & ~700K & - & Markdown Conversion (Pages) \\
\midrule
\textbf{SARD (Ours)} & \textbf{2025} & \textbf{Synthetic (Pages)} & \textbf{843.6K} & \textbf{689.7M} & \textbf{10} & \textbf{Large-scale Arabic book OCR} \\
\bottomrule
\end{tabular}
}
\end{table*}

Arabic is a globally prominent language, holding significant cultural and digital presence~\cite{marie2018arabicpsu}. Despite this, the development of large-scale, publicly available Arabic OCR datasets that accurately reflect the characteristics of printed books has not kept pace with the demands of state-of-the-art models. Many existing datasets concentrate on specific tasks, such as single-word recognition or handwritten text~\cite{berriche2024hybridpsu,ahmed2021novelpsu}. Datasets derived from scanned documents often suffer from inconsistent quality, noise, and distortions inherent in the scanning process. Furthermore, the manual annotation of large volumes of book pages is prohibitively expensive and time-consuming.

Arabic Optical Character Recognition (OCR) has long been an active research domain, driven by the pressing need to digitize large-scale collections of Arabic texts. Research efforts have predominantly focused on enhancing recognition accuracy for both printed and handwritten texts, acknowledging the unique attributes of the Arabic script, such as its cursive nature, context-dependent letter shapes, and prevalent ligatures. Datasets are paramount in this research, supplying the essential data for training and evaluating OCR models. 

The landscape of Arabic OCR datasets is diverse, encompassing various aspects of the script and different document types. Table \ref{tab:ocr_datasets_comp} offers a comparative overview of SARD alongside several prominent existing datasets, highlighting their principal characteristics. Existing Arabic OCR datasets can be broadly classified based on their text source and primary focus. Datasets focusing on printed text aim to replicate machine-generated typography. Examples include PATS-A01~\cite{PATS}, which provides line images from classical Arabic literature, and Yarmouk~\cite{Yarmouk}, containing diverse printed Arabic text. While valuable, these are often limited in scale or focus on isolated lines, not fully representing complex book layouts. 

Furthermore, APTI~\cite{mosbah2024adocrnet} is a large synthetic dataset for single-word recognition, and PAW~\cite{article_PAW} focuses on sub-word units. Handwritten text datasets like MADBase~\cite{ElSawy2016CNNFH}, ABAD~\cite{ICDAR}, OnlineKHATT~\cite{Online-KHATT}, and AlexU-Word~\cite{AlexU-Word} address variability in human handwriting but are not suitable for printed book OCR. Synthetic data generation, as seen in APTI~\cite{mosbah2024adocrnet}, IDPL-PFOD~\cite{farsitext} (Farsi), and Shotor~\cite{asadi2020shotor} (Farsi), helps overcome real-world data limitations. Arabic-Nougat~\cite{arabicnougats} converts scanned images to Markdown, focusing on output structure rather than providing raw training data for OCR models. Despite these efforts, a significant gap has persisted in large-scale, publicly available Arabic OCR datasets specifically simulating printed books with full page layouts, realistic structures, and extensive font diversity from a controlled synthetic process.

To address this critical data gap, we introduce SARD (Large-Scale Synthetic Arabic OCR Dataset). SARD is a novel, large-scale dataset composed entirely of synthetically generated Arabic document images, meticulously designed to mimic the structure and appearance of printed books. Employing a synthetic approach allows for the generation of a vast quantity of data with controlled variations in fonts, formatting, and content, While effectively mitigating noise, blurring, and other common artifacts associated with scanned document images. This controlled generation environment provides a clean, high-quality resource ideal for training robust OCR and vision-language models capable of generalizing well to diverse Arabic text styles and layouts. The dataset's structure and formatting are designed to simulate real-world book pages, making it particularly relevant for digitizing printed literature. 

As highlighted in Table \ref{tab:ocr_datasets_comp}, SARD distinguishes itself as a massive \emph{synthetic dataset} of \emph{full document pages}, with a deliberate focus on \emph{book-style formatting} and an unprecedented \emph{scale} and \emph{font diversity} (10 fonts compared to fewer or unspecified fonts in many other datasets) for this specific type of data. Its synthetic nature ensures clean, perfect ground truth and avoids the noise and distortions common in scanned documents, offering a controlled environment ideal for training robust models. This combination makes SARD uniquely suited for training and evaluating modern OCR and vision-language models designed to process structured documents effectively. The development of powerful deep learning models, including Vision-Language Models (VLMs)~\cite{najam2023analysis}, further necessitates large and diverse datasets like SARD to unlock their full potential for complex tasks such as digitizing extensive book collections with varied typography and layouts.

The main contributions of this work are:
\begin{itemize}
\item \textbf{Addressing a critical data gap} for Arabic OCR by introducing SARD, the first large-scale synthetic dataset specifically designed to simulate the structure of book-style documents. This fills a void for training modern, page-level OCR and vision-language models.
\item \textbf{Providing unprecedented scale and diversity} in a controlled environment, with SARD comprising over 840,000 document images and 690 million words, rendered across ten distinct Arabic fonts to capture significant typographic variation.
\item \textbf{Leveraging synthetic generation} to create a high-quality, noise-free dataset with perfect ground truth alignment, enabling scalability and precise control over layout parameters and text features (like diacritics and RTL flow) relevant to book transcription.
\item \textbf{Establishing SARD as a benchmark resource} for Arabic OCR by providing initial performance evaluations for a range of models, from traditional engines to state-of-the-art vision-language models, on challenging book-style layouts, thus setting baseline metrics (CER, WER, BLEU) for future research.
\item \textbf{Contributing to the research community} by publicly releasing the SARD dataset and its comprehensive generation scripts, fostering further advancements in robust Arabic OCR, document analysis, and vision-language modeling for complex printed materials.
\end{itemize}

\section{Results}

To demonstrate the utility of SARD for benchmarking and to provide initial performance metrics, we evaluated several representative OCR models. Our evaluation centered on standard OCR metrics: Character Error Rate (CER), Word Error Rate (WER), and Bilingual Evaluation Understudy (BLEU) score. Lower CER and WER values signify better accuracy at the character and word levels, respectively. A higher BLEU score indicates greater overall sequence similarity between the predicted and ground truth text, a metric commonly used in text generation tasks but also valuable here for assessing the fluency and overall correctness of the recognized text block.

We selected a diverse array of models for this evaluation. These included Tesseract \cite{smith2007overview}, a widely adopted traditional OCR engine with robust Arabic language support; EasyOCR \cite{JaidedAI_EasyOCR}, a popular deep learning-based OCR library supporting numerous languages, including Arabic; AIN~\cite{heakl2025ain}, an Arabic-specific OCR model fine-tuned vision-language model; Mistral OCR~\cite{mistralOCR2025}, a recent large vision-language model with multilingual capabilities; arabic-large-nougat~\cite{rashad2024arabic}, a specialized model for Arabic document understanding; and Qari-OCR V0.2 and V0.3~\footnote{\url{https://huggingface.co/collections/NAMAA-Space/qari-ocr-a-high-accuracy-model-for-arabic-optical-character-67c6cdff9584ef0684391335}}, a recent vision-language model specifically fine-tuned for Arabic OCR.

\begin{table}[t]
\centering
\caption{Word Error Rate (WER), with best results bolded.}
\label{tab:wer_table_avg_sorted} 
\resizebox{\columnwidth}{!}{%
\begin{tabular}{lrrrrrrrrrrr} 
\toprule
Font & Al-Jazeera & Amiri & Arial & Calibri & Jozoor & Lateef & Noto Naskh & Sakkal M. & Scheherazade & Thabit & Average \\ 
Model & & & & & Font & & Arabic UI & & & & \\ 
\midrule
Qari v0.2      & \textbf{0.1459} & 0.2670 & 0.3080 & 0.2490 & \textbf{0.0936} & \textbf{0.2239} & \textbf{0.2591} & 0.2930 & 0.2110 & \textbf{0.1433} & \textbf{0.2194} \\
Mistral-OCR    & 0.2957 & \textbf{0.0410} & \textbf{0.2480} & \textbf{0.1660} & 0.2940 & 0.3262 & 0.2878 & \textbf{0.1940} & \textbf{0.0990} & 0.2963 & 0.2248 \\
Tesseract      & 0.3429 & 0.4260 & 0.3120 & 0.3300 & 0.3300 & 0.3688 & 0.3640 & 0.3430 & 0.3310 & 0.3444 & 0.3492 \\
Qari v0.3      & 0.4004 & 0.3690 & 0.4820 & 0.4320 & 0.4116 & 0.4239 & 0.4845 & 0.4490 & 0.4640 & 0.3413 & 0.4258 \\
Arabic-Nougat  & 0.3993 & 0.4520 & 0.6780 & 0.5810 & 0.3736 & 0.4987 & 0.4884 & 0.4460 & 0.3850 & 0.3908 & 0.4693 \\
AIN            & 0.7589 & 0.8310 & 0.7300 & 0.7540 & 0.7317 & 0.6213 & 0.6755 & 0.7700 & 0.8370 & 0.7528 & 0.7462 \\
EasyOCR        & 0.7853 & 0.7160 & 0.7360 & 0.7280 & 0.7829 & 0.7766 & 0.7888 & 0.7630 & 0.7930 & 0.7858 & 0.7655 \\
\bottomrule
\end{tabular}%
}
\end{table}

\begin{table}[t]
\centering
\caption{Character Error Rate (CER) results.}
\label{tab:cer_table_avg_sorted} 
\resizebox{\columnwidth}{!}{%
\begin{tabular}{lrrrrrrrrrrr} 
\toprule
Font & Al-Jazeera & Amiri & Arial & Calibri & Jozoor & Lateef & Noto Naskh & Sakkal M. & Scheherazade & Thabit & Average \\ 
Model & & & & & Font & & Arabic UI & & & & \\ 
\midrule
Mistral-OCR    & \textbf{0.0679} & \textbf{0.0110} & \textbf{0.0510} & \textbf{0.0350} & 0.0705 & \textbf{0.0794} & \textbf{0.0650} & \textbf{0.0400} & \textbf{0.0200} & \textbf{0.0705} & \textbf{0.0510} \\
Qari v0.2      & 0.0924 & 0.2000 & 0.2300 & 0.1930 & \textbf{0.0490} & 0.1609 & 0.1884 & 0.2160 & 0.1560 & 0.0926 & 0.1578 \\
Tesseract      & 0.1526 & 0.2360 & 0.1400 & 0.1500 & 0.1392 & 0.1639 & 0.1755 & 0.1550 & 0.1540 & 0.1533 & 0.1620 \\ 
Arabic-Nougat  & 0.1988 & 0.2190 & 0.4090 & 0.3360 & 0.1746 & 0.2776 & 0.2635 & 0.2150 & 0.1750 & 0.1929 & 0.2461 \\ 
Qari v0.3      & 0.3638 & 0.3500 & 0.4610 & 0.4000 & 0.3915 & 0.3947 & 0.4463 & 0.4240 & 0.4830 & 0.3173 & 0.3932 \\ 
EasyOCR        & 0.6128 & 0.5570 & 0.5680 & 0.5480 & 0.6057 & 0.5839 & 0.6154 & 0.6060 & 0.6440 & 0.6151 & 0.5956 \\ 
AIN            & 0.7226 & 0.8040 & 0.6980 & 0.7230 & 0.6961 & 0.5377 & 0.6130 & 0.7310 & 0.8150 & 0.7154 & 0.7056 \\ 
\bottomrule
\end{tabular}%
}
\end{table}

The evaluation was conducted on a test set split from SARD. For the results presented in Tables~\ref{tab:wer_table_avg_sorted}, \ref{tab:cer_table_avg_sorted}, and \ref{tab:bleu_table_avg_sorted}, 200 randomly selected images were used for each of the ten fonts incorporated in the dataset, resulting in a total test set of 2000 images for this comprehensive analysis. These tables present the detailed WER, CER, and BLEU scores, respectively, for each evaluated model across all ten SARD fonts, with models ordered by their average performance for each metric.

The evaluation results reveal considerable variation in performance across the tested models and fonts, underscoring SARD's capability to differentiate model strengths and weaknesses.

\begin{table}[t]
\centering
\caption{BLEU Score results}
\label{tab:bleu_table_avg_sorted} 
\resizebox{\columnwidth}{!}{%
\begin{tabular}{lrrrrrrrrrrr} 
\toprule
Font & Al-Jazeera & Amiri & Arial & Calibri & Jozoor & Lateef & Noto Naskh & Sakkal M. & Scheherazade & Thabit & Average \\ 
Model & & & & & Font & & Arabic UI & & & & \\ 
\midrule
Qari v0.2      & \textbf{0.8116} & 0.7230 & \textbf{0.7030} & \textbf{0.7450} & \textbf{0.8346} & \textbf{0.7572} & \textbf{0.7443} & \textbf{0.7010} & 0.7820 & \textbf{0.8166} & \textbf{0.7618} \\ 
Mistral-OCR    & 0.5745 & \textbf{0.9200} & 0.6340 & 0.7460 & 0.5808 & 0.5305 & 0.5795 & 0.7150 & \textbf{0.8450} & 0.5751 & 0.6700 \\ 
Tesseract      & 0.5449 & 0.4240 & 0.5920 & 0.5570 & 0.5512 & 0.5360 & 0.5249 & 0.5190 & 0.5800 & 0.5443 & 0.5373 \\ 
Arabic-Nougat  & 0.4378 & 0.3550 & 0.1830 & 0.2710 & 0.4632 & 0.3567 & 0.3560 & 0.3670 & 0.4380 & 0.4409 & 0.3669 \\ 
Qari v0.3      & 0.3806 & 0.3460 & 0.2290 & 0.2860 & 0.3755 & 0.3337 & 0.2690 & 0.2790 & 0.2550 & 0.3779 & 0.3132 \\ 
EasyOCR        & 0.2588 & 0.3410 & 0.3410 & 0.3400 & 0.2604 & 0.2580 & 0.2602 & 0.2930 & 0.2960 & 0.2600 & 0.2908 \\ 
AIN            & 0.0741 & 0.0240 & 0.1030 & 0.0730 & 0.0976 & 0.3487 & 0.2016 & 0.0680 & 0.0390 & 0.0783 & 0.1107 \\ 
\bottomrule
\end{tabular}%
}
\end{table}

Across all ten fonts, advanced vision-language models demonstrated superior performance. As detailed in Table~\ref{tab:cer_table_avg_sorted}, Mistral OCR achieved the best average Character Error Rate (0.0510), showcasing remarkable precision at the character level. It secured the lowest CER on eight out of the ten fonts, with exceptionally low rates on Amiri (0.011), Scheherazade New (0.020), and Calibri (0.035).

Qari-OCR V0.2 emerged as the top performer in terms of average Word Error Rate (0.2194, Table~\ref{tab:wer_table_avg_sorted}) and average BLEU score (0.7618, Table~\ref{tab:bleu_table_avg_sorted}). This indicates its strength in producing correct words and overall fluent text sequences. It closely followed Mistral OCR in average CER (0.1578). Mistral OCR, in turn, had a highly competitive average WER of 0.2248, making these two models the clear front-runners.

Specifically, Qari-OCR V0.2 achieved the best WER on four fonts: Al-Jazeera (0.1459), Jozoor (0.0936), Lateef (0.2239), and Thabit (0.1433). Its dominance in BLEU scores was even more pronounced, securing the highest scores on eight out of ten fonts, signifying excellent overall text block recognition quality. Mistral OCR complemented this by achieving the best WER on the remaining six fonts, including an outstanding 0.0410 on Amiri and 0.0990 on Scheherazade New, and the highest BLEU scores on Amiri (0.9200) and Scheherazade New (0.8450).

Tesseract, the traditional OCR engine, delivered moderate and relatively consistent performance, ranking third on average for CER (0.1620) and WER (0.3492), and third for BLEU score (0.5373). While not matching the top deep learning models, its performance was notably better than several other deep learning-based approaches, particularly on fonts like Arial (CER 0.140, WER 0.312) and Calibri (CER 0.150, WER 0.330).

Arabic-Nougat, a specialized model for Arabic document understanding, exhibited varied performance across fonts. Its average CER (0.2461) and WER (0.4693) place it in the mid-to-lower tier. It performed relatively well on fonts like Jozoor (CER 0.1746, WER 0.3736) and Scheherazade New (CER 0.1750, WER 0.3850), but struggled with others like Arial (CER 0.4090, WER 0.6780).

Qari-OCR V0.3 showed a significant performance drop compared to its V0.2 counterpart across all metrics. With an average CER of 0.3932, WER of 0.4258, and BLEU of 0.3132, it generally performed worse than Tesseract and Arabic-Nougat, indicating potential issues in its fine-tuning or architecture for this diverse dataset.

EasyOCR, a general-purpose deep learning OCR library, and AIN, an Arabic-specific fine-tuned model, consistently recorded the highest error rates and lowest BLEU scores. EasyOCR's average CER was 0.5956 and WER was 0.7655. AIN performed even worse, with an average CER of 0.7056 and WER of 0.7462, and the lowest average BLEU score of 0.1107. This suggests that general-purpose libraries or specific fine-tuning approaches may require significant adaptation or more diverse training data to handle the typographic variety presented by SARD.

A crucial insight from this comprehensive benchmark is the significant performance variation across different fonts for nearly all models. For instance, Mistral OCR's CER on Amiri (0.0110) was exceptionally low, while its CER on Thabit (0.0705), though still the best for that font, was considerably higher. Similarly, Qari-OCR V0.2's best WER was on Jozoor (0.0936), but it was 0.3080 on Arial. Even Tesseract showed variability, with its WER on Arial (0.3120) being noticeably better than on Amiri (0.4260). This underscores the typographic challenge SARD introduces and its effectiveness in revealing the robustness and generalization capabilities of OCR models to font diversity, a critical aspect for real-world Arabic document processing.

Our benchmarks across the ten fonts in SARD effectively differentiate the capabilities of various OCR models for Arabic text. The results unequivocally demonstrate that recent, advanced vision-language models, particularly Mistral OCR and Qari-OCR V0.2, offer substantially superior accuracy and fluency for Arabic OCR tasks compared to traditional engines, general-purpose deep learning libraries, and some specialized models when tested on this typographically diverse dataset. The pronounced performance variance observed across different fonts for most models highlights the critical need for datasets like SARD, which systematically incorporate rich typographic diversity. This benchmark not only provides valuable initial performance metrics but also robustly validates SARD as an essential resource for the comprehensive evaluation, development, and advancement of Arabic OCR technologies.

\section{Discussion}
The evaluation results offer significant insights into the capabilities of various OCR methodologies when applied to structured, multi-font Arabic book pages as simulated by SARD. A key observation is the substantial performance disparity among the tested models, underscoring the demanding nature of book-style Arabic OCR and illuminating crucial differences between traditional OCR engines, general-purpose deep learning OCR libraries, and advanced Vision-Language Models (VLMs) or specialized architectures.

General-purpose tools like EasyOCR, despite being deep learning-based, exhibited high error rates (CER often $>$ 50\%, WER $>$ 70\%) and low BLEU scores on SARD. This suggests that standard, out-of-the-box deep learning OCR implementations, without targeted training or fine-tuning for this specific document style and typographic diversity, struggle with the complexities SARD presents. Conversely, Tesseract, a mature traditional OCR engine, demonstrated markedly lower error rates and higher BLEU scores than EasyOCR. This indicates that Tesseract's refined text layout analysis and recognition pipelines, developed over many years, can offer superior performance on structured printed Arabic text compared to some unadapted general deep learning approaches.

Models employing more sophisticated architectures, particularly VLMs and those specialized for document understanding, generally yielded better results, though with notable variance. Mistral OCR, a powerful multilingual VLM, established a high benchmark, achieving remarkably low CER (e.g., 1.1\% on Amiri) and WER, alongside high BLEU scores. This strong performance suggests that comprehensive VLMs possess potent capabilities for concurrently processing visual layout and intricate Arabic script features found in SARD, showcasing their potential for high-accuracy book OCR. Qari OCR, a VLM fine-tuned for Arabic OCR, also significantly outperformed Tesseract and EasyOCR, confirming the benefits of task-specific VLM adaptation for Arabic document transcription. The Arabic-large-nougat model, designed for Arabic document image understanding, showed inconsistent performance across fonts, indicating challenges in maintaining uniform accuracy over diverse typographies even with specialized architectures.

Unexpectedly, AIN, described as an Arabic-specific, fine-tuned VLM, registered the highest error rates and lowest BLEU scores on SARD, performing worse than the traditional Tesseract engine. This outcome underscores that VLM architecture or fine-tuning alone does not guarantee superior performance on a diverse, structured dataset like SARD. Factors such as the base VLM, the nature and scale of fine-tuning data, or implementation specifics can heavily influence efficacy when confronted with SARD's unique combination of synthetic book-style data and font variety.

The clear font-dependent performance variations observed for most models (e.g., Mistral's higher error rates on Arial and Sakkal Majalla versus Amiri) reinforce the value of SARD's inclusion of ten diverse fonts. Training and evaluating on such a typographically rich dataset is crucial for developing systems that can robustly generalize across the wide spectrum of Arabic printed styles.

These findings establish initial baselines on SARD and affirm its utility as a challenging and valuable dataset for assessing Arabic OCR systems, especially those targeting book-style documents. The significant performance gaps highlight SARD's discriminatory power as a benchmark for driving advancements in Arabic OCR for complex layouts and diverse typography. Achieving consistently high accuracy across all font styles and book-like layouts, essential for large-scale digitization, will necessitate further research. This may involve more extensive or tailored pre-training and fine-tuning of VLMs or the development of novel architectures specifically adapted to the complexities highlighted by datasets like SARD.

While SARD represents a significant step forward, certain limitations should be acknowledged. As a purely synthetic dataset, SARD does not incorporate real-world document degradations such as uneven lighting, paper texture, ink bleed-through, or page skew, which can impact model performance on actual scanned materials. Although it includes ten diverse Arabic fonts, this is a subset of the vast typographic landscape of Arabic script, especially concerning historical or calligraphic styles. The dataset also simplifies document structure by omitting complex elements like multi-column layouts, embedded tables, figures, and footnotes often found in academic or intricate publications. Furthermore, SARD focuses exclusively on printed text, not addressing handwritten content, and may not fully represent specialized vocabularies from highly technical or niche domains.

Future work will aim to address these limitations. Planned enhancements include integrating synthetic noise models to simulate common scanning artifacts, thereby bridging the gap between synthetic and real-world data. We intend to expand font coverage to include a broader range of historical, calligraphic, and contemporary styles. Generating more complex document layouts, incorporating elements like tables, multi-column text, and marginalia, is also a priority. Exploring the development of a companion dataset for handwritten Arabic text could complement SARD. Additionally, creating domain-specific extensions of SARD, focusing on technical or specialized vocabularies, and incorporating mixed-script content (e.g., Arabic with Latin script for citations or technical terms) would further enhance its applicability. These improvements are envisioned to strengthen SARD's role as a comprehensive and evolving resource for Arabic OCR research.

The introduction of SARD provides a much-needed large-scale, high-quality resource for advancing Arabic OCR. The initial benchmarks confirm its challenging nature and its potential to drive innovation in models capable of handling the complexities of Arabic book-style text.
\begin{figure}[t]
 \centering
  \includegraphics[width=\textwidth]{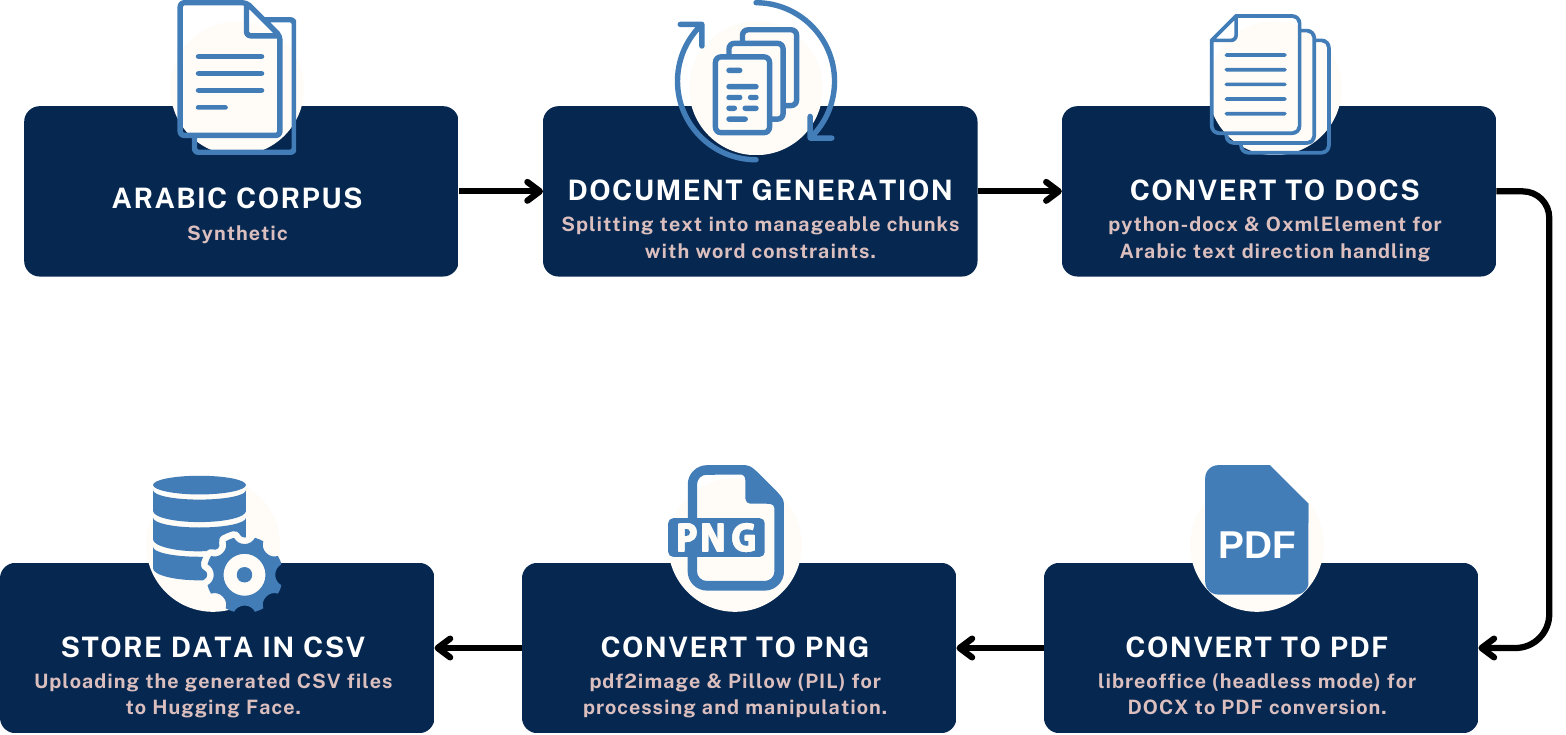}
  \caption{SARD Corpus Generation Pipeline}
  \label{fig:1}
\end{figure}

\section{Methods}
The foundation of SARD is an extensive corpus of Arabic text sourced from diverse domains. To ensure linguistic richness and cover a broad spectrum of vocabulary and writing styles pertinent to printed literature, we utilized text from numerous articles spanning categories such as Culture, Fatawa \& Counsels, Literature \& Language, Bibliography, Publications \& Competitions, Shariah, Social, Translations, and News. Our collection comprises over 133,000 unique articles, with the distribution across these categories detailed in Table~\ref{tab:article_distribution}. As indicated, substantial contributions originate from Shariah, Bibliography, and News, complemented by significant content from other varied fields. This comprehensive collection ensures that models trained on SARD encounter a wide range of topics, vocabulary, and sentence structures, thereby improving their generalization capabilities across different textual domains commonly found in books.

\begin{table}[t]
\centering
\renewcommand{\arraystretch}{1.2}
\caption{Distribution of source articles across different domains used for SARD.}
\label{tab:article_distribution}
\begin{tabular}{ll}
\toprule
\textbf{Category} & \textbf{No. of Articles} \\
\midrule
Culture & 13,253 \\
Fatawa \& Counsels & 8,096 \\
Literature \& Language & 11,581 \\
Bibliography & 26,393 \\
Publications \& Competitions & 1,123 \\
Shariah & 46,665 \\
Social & 8,827 \\
Translations & 443 \\
News & 16,725 \\
\midrule
\textbf{Total Articles} & \textbf{133,105} \\
\bottomrule
\end{tabular}
\end{table}

A critical aspect of recognizing book-style OCR is the ability to handle diverse typographic styles. To this end, SARD is generated using ten distinct Arabic fonts that represent a broad spectrum of styles commonly encountered in Arabic publications. The selection includes Amiri (a classical Naskh typeface inspired by early 20th-century typography), Arial and Calibri (modern sans-serif fonts prevalent in digital publications), Sakkal Majalla (widely used in contemporary Arabic publishing), Scheherazade New (a traditional-style font based on classical manuscript aesthetics), Noto Naskh Arabic UI (part of Google's Noto font family, designed for clear UI rendering), Lateef (optimized for readability at small sizes), Thabit (a monospaced font often used for technical documentation), Jozoor Font (featuring more stylized character forms), and Al-Jazeera-Arabic-Regular (representing modern journalistic typography). This carefully curated selection provides extensive typographic variation while covering many prevalent styles in printed and digital Arabic literature.

As illustrated in Figure~\ref{fig:1}, SARD is generated through a multi-step synthetic pipeline designed to convert raw text into realistic-looking document images while maintaining control over formatting and quality. The pipeline commences with text selection and preparation, where raw text articles from the chosen domains are cleaned and readied for rendering. This crucial initial stage involves addressing Arabic encoding issues, removing unwanted characters, normalizing diacritics, and structuring the text into logical units such as paragraphs. Once prepared, the text undergoes layout and formatting according to standard book layout specifications. Page dimensions, margins, font styles, and font sizes are meticulously configured at this stage.

To introduce variability and simulate different book styles, we employ the font specifications detailed in Table~\ref{tab:font_specs}, which include a range of words per page and specific font sizes for each typeface. This approach ensures that the generated pages exhibit realistic line breaks, paragraph spacing, and text density, mirroring those found in actual printed books. The overall page dimensions, margins, and structural characteristics are defined by the parameters listed in Table~\ref{tab:layout_params}. These specifications, common in publishing, contribute significantly to the realism of the synthetic data by simulating standard book layouts.

\begin{table}[ht]
\centering
\caption{Font specifications used in SARD generation}
\label{tab:font_specs}
\small
\begin{tabular}{@{}lcc@{}}
\toprule
\textbf{Font} & \textbf{Words/Page Range} & \textbf{Font Size} \\
\midrule
Sakkal Majalla & 50–300 & 14 pt \\
Arial & 50–500 & 12 pt \\
Calibri & 50–500 & 12 pt \\
Amiri & 50–300 & 12 pt \\
Scheherazade New & 50–250 & 12 pt \\
Noto Naskh Arabic UI & 50–400 & 12 pt \\
Lateef & 50–350 & 14 pt \\
Al-Jazeera-Arabic & 50–250 & 12 pt \\
Thabit & 50–240 & 12 pt \\
Jozoor Font & 50–200 & 12 pt \\
\bottomrule
\end{tabular}
\end{table}

\begin{table}[ht]
\centering
\caption{Page layout parameters used in SARD generation}
\label{tab:layout_params}
\small
\begin{tabular}{@{}lc@{}}
\toprule
\textbf{Specification} & \textbf{Value} \\
\midrule
Page Size & A4 (8.27 × 11.69 in) \\
Left Margin & 0.9 in \\
Right Margin & 0.9 in \\
Top Margin & 1.0 in \\
Bottom Margin & 1.0 in \\
Gutter Margin & 0.2 in \\
Resolution & 300 DPI \\
Color Mode & Grayscale \\
Page Direction & Right-to-Left \\
Text Alignment & Right \\
Line Spacing & 1.15 \\
\bottomrule
\end{tabular}
\end{table}

The subsequent stage involves image rendering, where each document page is rendered as a high-resolution image (300 DPI) using the specified layout and fonts. This process precisely controls the visual appearance of the text, including character shapes, spacing, and baseline alignment, according to the chosen font and size. We employ anti-aliasing techniques to ensure high-quality rendering that closely mimics professional printing quality.

The final critical stage is ground truth generation. Concurrently with each image, the exact source text and formatting parameters used in its generation are saved as metadata. This pairing of image with ground truth text is fundamental for training supervised OCR models and provides a perfect alignment between visual content and textual data, devoid of the noise or inconsistencies typically found in manually annotated datasets.

\begin{figure}[htbp] 
\centering
\includegraphics[width=\textwidth]{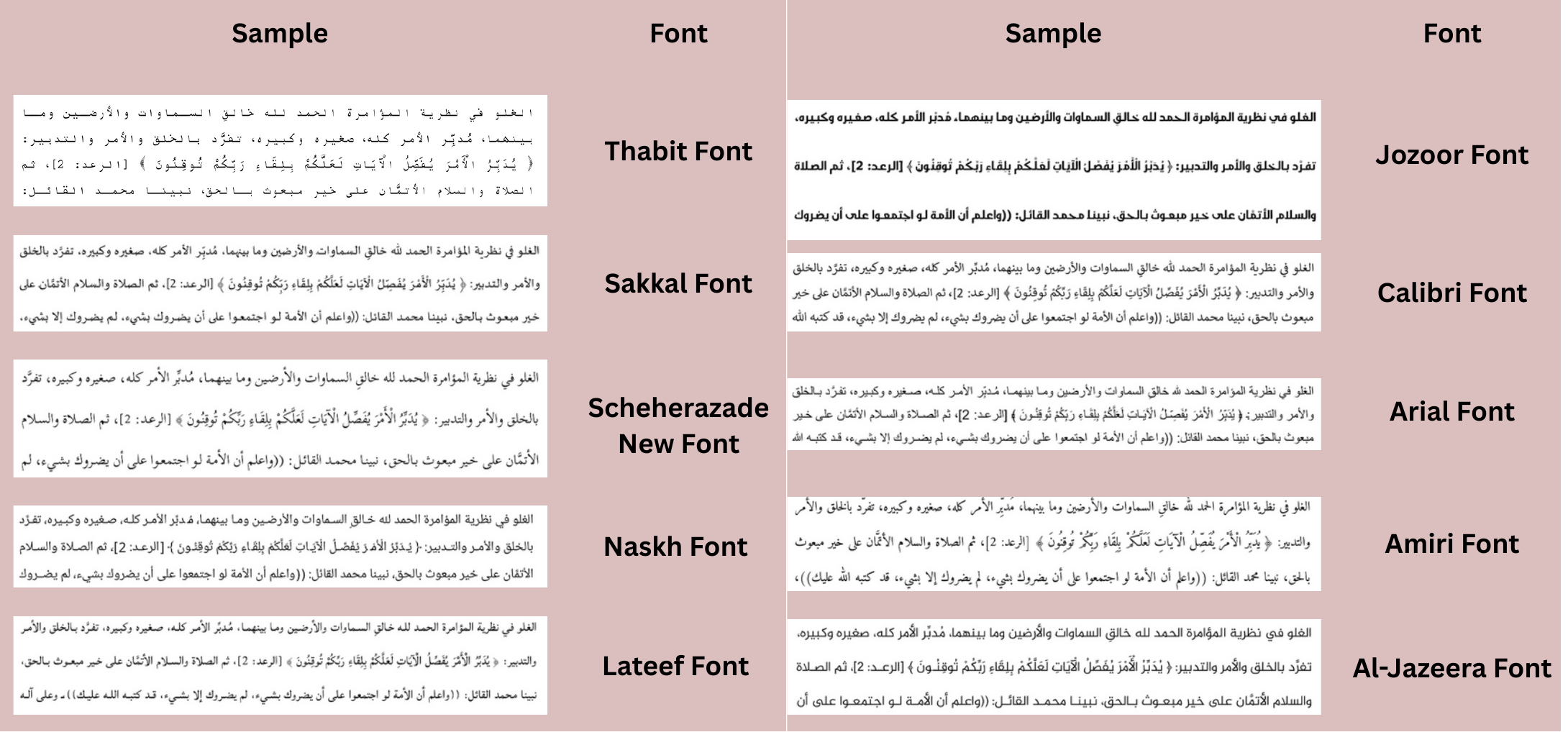}
\caption{Samples of the Arabic fonts included in SARD}
\label{fig:sard_font_showcase}
\end{figure}

The synthetic nature of SARD ensures that the ground truth text perfectly matches the generated image content, eliminating recognition errors commonly found in manually transcribed datasets. Furthermore, by controlling factors such as font type, size, and layout, the dataset is designed to systematically explore the variability relevant to book OCR tasks. Figure~\ref{fig:sard_font_showcase} illustrates this diversity by presenting samples of each font included in SARD, with each font style showcased across three lines of text.

This systematic methodology facilitates the creation of a massive, high-quality dataset tailored for training and evaluating OCR models on Arabic book-style documents, addressing the specific challenges posed by diverse fonts and realistic layouts.

\section*{Conclusion}
We introduced SARD, a large-scale synthetic Arabic OCR dataset specifically designed for book-style text recognition. Comprising 843,622 document images with 690 million words across ten diverse fonts, SARD addresses the critical need for comprehensive training data in Arabic OCR research. The dataset's synthetic nature ensures perfect ground truth alignment while enabling controlled typographic and structural variation essential for robust model development.

Our benchmark evaluations across multiple OCR approaches revealed significant performance variations, with advanced vision-language models demonstrating promising results while highlighting the continued challenges faced by general-purpose OCR tools on structured Arabic text. By publicly releasing SARD and its generation scripts, we aim to accelerate research in Arabic document understanding and enable more effective digitization of Arabic literary heritage. The dataset provides a valuable foundation for future innovations in multilingual OCR and document analysis systems capable of handling the rich complexity of Arabic typography and book layouts.

\section*{Data Availability}
The SARD dataset is freely accessible from our Hugging Face repository:
\begin{center}
  \url{https://huggingface.co/collections/riotu-lab/sard-large-scale-synthetic-arabic-ocr-dataset-68204eae3ccd1e0216fb68ca}
\end{center}
We encourage the research community to utilize SARD in their work.

\section*{Code Availability}
The generation scripts used for creating the SARD dataset are also publicly available to promote transparency and enable further development:
\begin{center}
  \url{https://github.com/riotu-lab/text2image}
\end{center}
We hope these resources contribute to the advancement of Arabic OCR technology.

\section*{Acknowledgments}
The authors would like to acknowledge the support of Prince Sultan University for paying the Article Processing Charges (APC) of this publication.

\bibliographystyle{elsarticle-num}
\bibliography{custom} 

\end{document}